\documentclass[review]{elsarticle}

\journal{Journal of \LaTeX\ Templates}

\usepackage[numbers]{natbib}
\usepackage{graphicx}
\usepackage{multirow}

\usepackage{caption} 
\usepackage{bm}
\usepackage{amssymb}
\usepackage[ruled]{algorithm2e}
\usepackage{booktabs}

\begin{document}

\begin{frontmatter}

\title{Multi-view Hierarchical Clustering}

\author[mymainaddress]{Qinghai Zheng}

\author[mymainaddress]{Jihua Zhu}
\cortext[mycorrespondingauthor]{Corresponding author}
\ead{zhujh@xjtu.edu.cn}

\author[mymainaddress]{Shuangxun Ma}

\address[mymainaddress]{School of Software Engineering, Xi'an Jiaotong University, Xi'an 710049, China}

\begin{abstract}
This paper focuses on the multi-view clustering, which aims to promote clustering results with multi-view data. Usually, most existing works suffer from the issues of parameter selection and high computational complexity. To overcome these limitations, we propose a Multi-view Hierarchical Clustering (MHC), which partitions multi-view data recursively at multiple levels of granularity. Specifically, MHC consists of two important components: the cosine distance integration (CDI) and the nearest neighbor agglomeration (NNA). The CDI can explore the underlying complementary information of multi-view data so as to learn an essential distance matrix, which is utilized in NNA to obtain the clustering results. Significantly, the proposed MHC can be easily and effectively employed in real-world applications without parameter selection. Experiments on nine benchmark datasets illustrate the superiority of our method comparing to several state-of-the-art multi-view clustering methods.
\end{abstract}


\end{frontmatter}

\section{Introduction}

As technology evolves, data collected from multiple sources or various measurements are common in practice \cite{xu2013survey}. Different from single-view data, more information can be contained in multi-view data \cite{sun2013survey}. For example, an image can be comprehensive described by multiple features, including 
SIFT, GIST, LBP, etc. As an important learning paradigm in artificial intelligence, multi-view clustering attracts considerable attention in recent years \cite{li2018survey,chao2017survey,zhao2017multisurvey}. By investigating the underlying consensus and complementary information of multi-view data, numerous multi-view clustering methods are proposed and varied in the way of multi-view information exploration \cite{xia2014robust,cao2015diversity,zhang2017latent,xie2018unifying,zhang2018binary,peng2019comic,kang2020large}.

\begin{figure*}
	\begin{center}
		\includegraphics[width=1\linewidth]{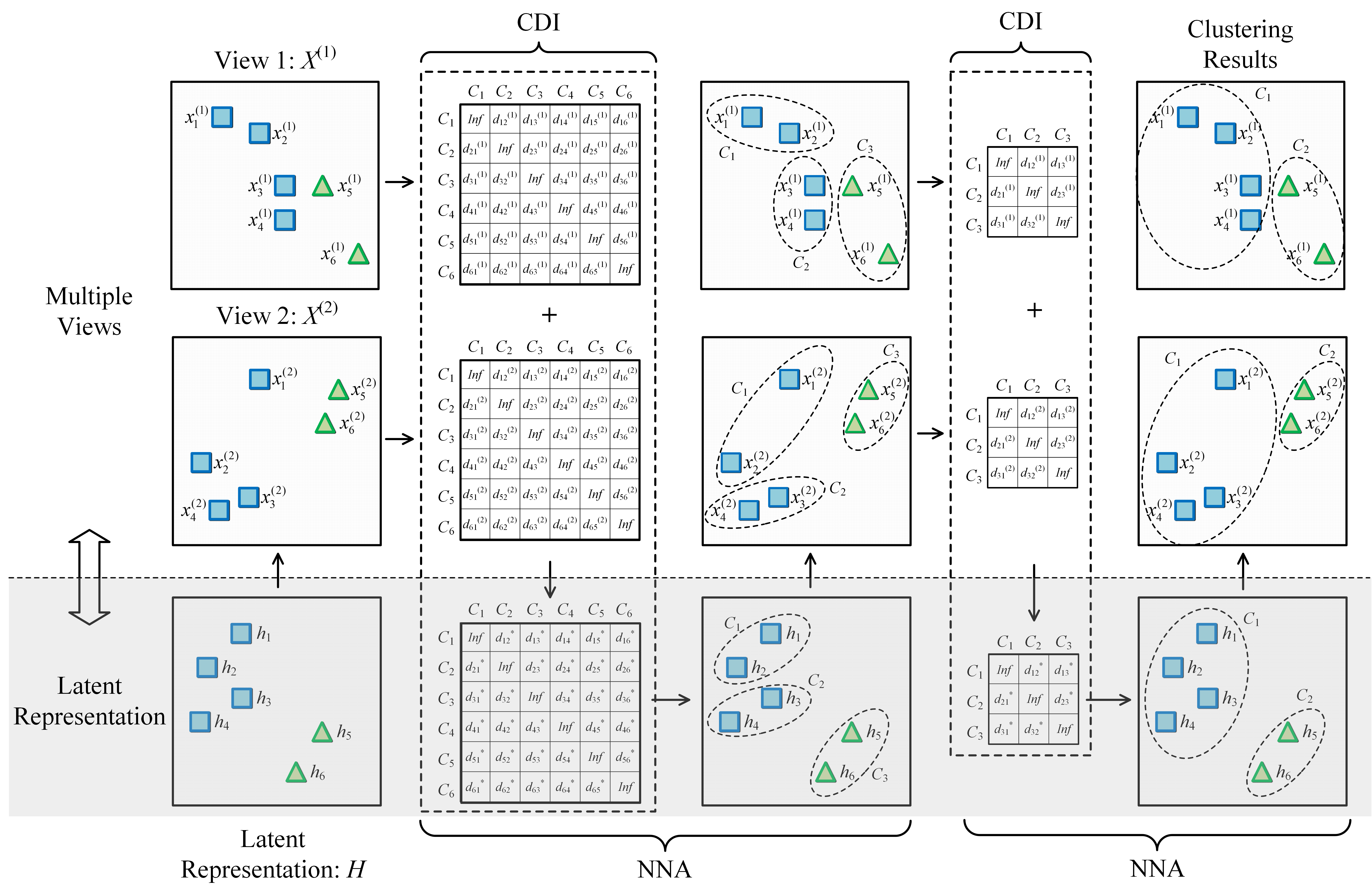}
	\end{center}
	\caption{The flowchart of the proposed Multi-view Hierarchical Clustering (MHC). We take multi-view data with two views, i.e., $\{{\bm{X}}^{(1)},{\bm{X}}^{(2)}\}$, for example. Specifically, we assume that multiple views can be generated from the latent representation ${\bm{H}}$. In each iteration of our method, the cosine distance integration (CDI) learns an essential distance matrix with the complementary information of multiple views, then the nearest neighbor agglomeration (NNA) merges the nearest clusters into a new cluster according to the learned distance matrix. As will be discussed later, we calculate the underlying distance of the $a$ and $b$ clusters with multiple views by $d_{ab}^{*}=\frac{1}{2}(d_{ab}^{(1)}+d_{ab}^{(2)})$, which contains the complementary information of multi-view data.}
	\label{Flowchart_MHC}
\end{figure*}

Although the promising performance can be achieved by multi-view clustering, most existing methods suffer from the following two issues in multi-view clustering: a) parameter selection, and b) high computational complexity. In most existing methods, such as the multi-view spectral clustering and multi-view subspace clustering, parameter selection is an indispensable part \cite{wu2019essential,zheng2020feature}. However, prior information, such as label information, is required to guide this specific process for parameter selection, which is difficult for multi-view clustering \cite{peng2019comic,zhang2020tensorized}. Besides, most existing multi-view clustering methods have high computational complexity, for example, computational complexities of both multi-view spectral clustering and multi-view subspace clustering are ${\cal O}(n^3)$, where $n$ is the number of data samples. Obviously, these aforementioned limitations seriously hinder the practicability and development of multi-view clustering \cite{peng2019comic,kang2020large}. Actually, some works are conducted to obtain large-scale multi-view clustering in recent years, such as the binary multi-view clustering \cite{zhang2018binary}, and large-scale multi-view subspace clustering \cite{kang2020large}, but they still suffer from the issue of parameter selection. Therefore, how to cluster multi-view data effectively and without parameter selection is still a challenging and important task in practice. 

In this paper, we propose a novel Multi-view Hierarchical Clustering (MHC) to overcome above issues. Our MHC may probably be the first multi-view clustering method based on the hierarchical clustering, which partitions data recursively and provides clustering results at an increasingly finer granularity \cite{reddy2013survey,charikar2017approximate,moseley2017approximation,sarfraz2019efficient,cohen2019hierarchical,xie2020hierarchical}. The goal of MHC is to get promising clustering results by utilizing the multi-view information without parameter selection and with low computational complexity. To be specific, two important components are contained in the proposed method: the cosine distance integration (CDI) and the nearest neighbor agglomeration (NNA). The CDI aims to learn an essential distance matrix, which contains the complementary information of multi-view data, for clustering. It is worth noting that the CDI assumes that multiple views can be reconstructed from one fundamental latent representation \cite{zhang2017latent}. Regarding to NNA, it merges data samples with their nearest neighbor into new clusters according to the learned essential distance matrix in CDI. The NNA obeys the intuitive observation that one data sample with its nearest neighbor should be grouped into the same cluster in the clustering process \cite{zhang2019neural,sarfraz2019efficient,xie2020hierarchical}. The flowchart of our MHC is depicted in Fig.~\ref{Flowchart_MHC}. It is clear that the proposed MHC provides clustering results at an increasingly finer granularity, i.e., a hierarchy tree. To attain clustering results with the desired number of clusters, we can refine the closest division with larger number of clusters directly by merging two clusters at each iteration. The whole clustering process is free of the parameter selection and with ${\cal O}(n{\rm{log}}n)$ computational complexity, which will be analyzed in detail later. 

We summarize the main contributions of this paper as follows:
\begin{itemize}
	\item[1)] It proposes a Multi-view Hierarchical Clustering (MHC), which can obtain promising clustering results without parameter selection and with low computational complexity.
	\item[2)] By introducing the cosine distance integration and nearest neighbor agglomeration, the complementary information of multi-view data can be fully explored for clustering.
	\item[3)] Experimental results on nine real-world datasets illustrate the superiority of the proposed method.
\end{itemize}

\section{Related Works}
In this section, we will introduce the hierarchical clustering and multi-view clustering briefly.

\subsection{Hierarchical Clustering}
Hierarchical clustering is a clustering method which clusters data samples at multiple levels of granularity \cite{reddy2013survey,sarfraz2019efficient,cohen2019hierarchical,xie2020hierarchical}. Single-linkage, average-linkage, and complete-linkage are widely used agglomerative techniques for clustering. Most existing hierarchical clustering methods are designed for single-view data. Differently, the proposed MHC is probably the first hierarchical clustering method for multi-view data. Furthermore, our method utilizes the nearest neighbor instead of single-linkage, average-linkage, or complete-linkage, which can reduce the computational complexity considerably compared to traditional hierarchical clustering methods.

\subsection{Multi-view Clustering}
Many works are conduced to solve the multi-view clustering problem \cite{xia2014robust,chao2017survey,zhang2017latent,zhang2018binary,wang2019gmc,peng2019comic,zheng2020constrained}. According to the way of exploring underlying information among multiple views, we divide most existing multi-view clustering methods into three categories roughly: a) multi-view spectral clustering methods, b) multi-view subspace clustering methods, and c) other multi-view clustering methods. 

For multi-view spectral clustering, the core idea is to learn an intrinsic graph, which contains the information of multi-view data, then get clustering results by performing the spectral clustering algorithm on the learned graph. For example, the method proposed in \cite{xia2014robust} applies low-rank and sparse decomposition on probability matrices of different views to learn a shared probability matrix for clustering. By assigning weights to different graphs of multiple views, the method proposed in \cite{wang2019gmc} learns an unified graph in the mutual reinforcement way to improve clustering results. The method proposed in \cite{Ma2020MSCCI} takes the diversity of different views into consideration and builds a similarity matrix by investigating the complementary information of multi-view data for clustering.

Regarding to multi-view subspace clustering methods, the commonality is to learn a unified subspace representation of different views based on \cite{liu2013robust} or \cite{elhamifar2013sparse}. For example, the method proposed in \cite{zheng2020feature} introduces the view-specific corruption as well as sample-specific corruption to learn an intrinsic subspace representation by the concatenated features. Considering the complementary information of different views, the method proposed in \cite{zhang2017latent} learns a latent representation of multiple views and obtains subspace representations for clustering simultaneously. By introducing the bilinear factorization into the subspace representation learning, the consensus and complementary information of multi-view data are explored in \cite{zheng2020constrained}, and promising clustering results are achieved as well.

Besides, there are some other multi-view clustering methods proposed in recent years \cite{andrew2013deep,zhang2018binary,zhao2017NMFMSC,peng2019comic}. For example, the method proposed in \cite{zhang2018binary} introduces a binary based framework for clustering, which is able to deal with large-scale multi-view data in linear time. Considering the geometric consistency and cluster assignment consistency, the method proposed in \cite{peng2019comic} gets clustering results automatically. 
By leveraging the deep matrix factorization , the method proposed in \cite{zhao2017NMFMSC} fully explores the complementary information of multi-view data to benefit the clustering process.

\section{The Proposed Approach}
Given multi-view data $\{\bm{X}^{(i)}\}_{i=1}^v$ collected from $v$ views $m$ clusters, we aim to get clustering results without parameter selection and with low computational complexity. To this end, a novel Multi-view Hierarchical Clustering (MHC) is proposed in this paper. As can be observed in Fig.~\ref{Flowchart_MHC}, MHC consists of two important components in each iteration, i.e., the cosine distance integration (CDI) and the nearest neighbor agglomeration (NNA). To be clear, main symbols used in this paper are summarized in Table~\ref{Main_symbols}.

\begin{table}
	\caption{Main symbols.}\smallskip
	\label{Main_symbols}
	\centering
	\resizebox{0.85\columnwidth}{!}{
		\begin{tabular}{l|l}
			\toprule
			Symbol & Meaning\\
			\midrule
			$\bm{X}^{(i)}$ & The $i$-th view of multi-view data. \\
			\midrule
			$\bm{H}$ & Latent representation of multiple views. \\
			\midrule
			$\bm{x}_a^{(i)}$ & Sample $a$ in the $i$-th view. \\
			\midrule
			$\bm{h}_a^{(i)}$ & Sample $a$ in latent representation. \\
			\midrule
			$n$ & The number of data samples. \\
			\midrule
			$v$ & The number of views. \\
			\midrule
			$m$ & The number of clusters. \\
			\midrule
			$\rm{dim}_i$ & The dimensionality of the $i$-th view.\\
			\midrule
			$d_{ab}^{(i)}$ & Cosine distance of $a$ and $b$ samples in the $i$-th view.\\
			\midrule
			$d_{ab}^{*}$ & Cosine distance of $a$ and $b$ samples in latent representation.\\
			\bottomrule
	\end{tabular}}
\end{table}

\subsection{Cosine Distance Integration}
Regarding to the $i$-th view, $\bm{X}^{(i)} \in {{\mathbb{R}}^{{\rm{dim}_i} \times n}}$, where $n$ and ${\rm{dim}_i}$ are the number of data samples and the dimensionality of the $i$-th view  respectively. Let us denote $\bm{H}$ as the latent representation of multiple views, we assume that $\{\bm{X}^{(i)}\}_{i=1}^v$ can be reconstructed by $\bm{H}$, which contains the complementary information and reveals the underlying clustering structures. To be specific, relationships between the latent representation and multiple views can be formulated as follows:
\begin{equation}
{\bm{X}^{(i)}} = {\bm{P}^{(i)}}\bm{H},{\kern 1pt} {\kern 1pt} {\kern 1pt}{\bm{P}^{(i)}}^T{\bm{P}^{(i)}} = \bm{I},
\label{MappingHtoX}
\end{equation}
where ${\bm{P}^{(i)}}$ is an orthogonal matrix of the $i$-th view.

We adopt the cosine distance in this paper, and the cosine distance matrix of $a$ and $b$ samples in the $i$-th view can be constructed as follows:
\begin{equation}
d_{ab}^{(i)} = 1 - \frac{{\bm{x}_a^{{{(i)}^T}}\bm{x}_b^{(i)}}}{{\sqrt {\bm{x}_a^{{{(i)}^T}}\bm{x}_a^{(i)}} \sqrt {\bm{x}_b^{{{(i)}^T}}\bm{x}_b^{(i)}} }}.
\label{CosineDistance_Xi}
\end{equation}
As for the cosine distance matrix of $a$ and $b$ samples in latent representation, we have the following formulation:
\begin{equation}
d_{ab}^ *  = 1 - \frac{{h_a^{{T}}h_b}}{{\sqrt {h_a^{{T}}h_a} \sqrt {h_b^{{T}}h_b} }}.
\end{equation}
Substituting Eq.~(\ref{MappingHtoX}), i.e., $\bm{x}_a^{(i)}=\bm{P}^{(i)}\bm{h}_a$ and $\bm{x}_b^{(i)}=\bm{P}^{(i)}\bm{h}_b$, into Eq.~(\ref{CosineDistance_Xi}), the following formulation can be written:
\begin{equation}
\begin{array}{l}
d_{ab}^{(i)} = 1 - \frac{{{{({\bm{P}^{(i)}}{\bm{h}_a})}^T}({\bm{P}^{(i)}}{\bm{h}_b})}}{{\sqrt {{{({\bm{P}^{(i)}}{\bm{h}_a})}^T}({\bm{P}^{(i)}}{\bm{h}_a})} \sqrt {{{({\bm{P}^{(i)}}{\bm{h}_b})}^T}({\bm{P}^{(i)}}{\bm{h}_b})} }}\\
~~~~~{\kern 1pt}{\kern 1pt}= 1 - \frac{{(\bm{h}_a^T{\bm{P}^{{{(i)}^T}}}){\bm{P}^{(i)}}{\bm{h}_b}}}{{\sqrt {(\bm{h}_a^T{\bm{P}^{{{(i)}^T}}}){\bm{P}^{(i)}}{\bm{h}_a}} \sqrt {(\bm{h}_b^T{\bm{P}^{{{(i)}^T}}}){\bm{P}^{(i)}}{\bm{h}_b}} }}\\
~~~~~{\kern 1pt}{\kern 1pt}= 1 - \frac{{\bm{h}_a^T({\bm{P}^{{{(i)}^T}}}{\bm{P}^{(i)}}){\bm{h}_b}}}{{\sqrt {\bm{h}_a^T({\bm{P}^{{{(i)}^T}}}{\bm{P}^{(i)}}){\bm{h}_a}} \sqrt {\bm{h}_b^T({\bm{P}^{{{(i)}^T}}}{\bm{P}^{(i)}}){\bm{h}_b}} }}\\
~~~~~{\kern 1pt}{\kern 1pt}= 1 - \frac{{\bm{h}_a^T{\bm{h}_b}}}{{\sqrt {\bm{h}_a^T{\bm{h}_a}} \sqrt {\bm{h}_b^T{\bm{h}_b}} }}\\
~~~~~{\kern 1pt}{\kern 1pt} = d_{ab}^ * .
\end{array}
\end{equation}
Therefore, we can get the essential cosine distance of $a$ and $b$ samples in latent representation as follows:
\begin{equation}
d_{ab}^ *  = \frac{1}{v}\sum\limits_i^v {d_{ab}^{(i)}},
\label{Distance_Fusion}
\end{equation}
in which the complementary information is considered. We denote $\bm{D}^{(i)}$ as the cosine distance matrix of the $i$-the view, and let diagonal elements of $\bm{D}^{(i)}$ be infinity for simplicity. The essential cosine distance matrix of latent representation, i.e., $\bm{D}^ *$, can be achieved as follows:
\begin{equation}
\bm{D}^ *  = \frac{1}{v}\sum\limits_i^v {\bm{D}^{(i)}}.
\label{Get_Intrinsic_Distance_Matrix}
\end{equation}

Based on the learned essential cosine distance matrix $\bm{D}^ *$, the nearest neighbor agglomeration can be performed to get clustering results in current iteration.

\begin{algorithm}
	\caption{Algorithm of the proposed MHC}
	\label{Algorithm1}
	\LinesNumbered 
	\KwIn{Multi-view data, i.e., $\{ {{\bm{X}}^{(i)}}\} _{i = 1}^v$.}
	\KwOut{Clustering results at multiple levels of granularity, ${\cal{R}} = \{R_0, R_1, R_2, \cdots\}$.}
	\textbf{Clustering Process:}\\
	\tcp{Performing the CDI}
	\For{$i=1:v$}{
		Calculating distance matrix ${\bm{D}}^{(i)}$ by Eq.~(\ref{CosineDistance_Xi}),\\
	}
	Getting essential distance matrix ${\bm{D}}^{*}$ by Eq.~(\ref{Get_Intrinsic_Distance_Matrix}).\\
	\tcp{Performing the NNA}
	Constructing the graph $\cal{G}$ based on ${\bm{D}}^{*}$,\\
	Getting clustering results, i.e., $R_0$, based on current graph $\cal{G}$,\\
	Transferring current results $R_0$ to multiple views, and calculating new data samples as new input data by averaging data samples in the same cluster.\\
	\tcp{Performing the CDI and NNA recursively}		
	\While{current results have more than two clusters}{
		\tcp{Performing the CDI}
		\For{$i=1:v$}{
			Calculating distance matrix ${\bm{D}}_k^{(i)}$ based on new input data by Eq.~(\ref{CosineDistance_Xi}),\\
		}
		Getting essential distance matrix ${\bm{D}}_k^{*}$ by Eq.~(\ref{Get_Intrinsic_Distance_Matrix}).\\
		\tcp{Performing the NNA}
		Constructing the graph ${\cal{G}}_k$ based on ${\bm{D}}_k^{*}$,\\
		Getting clustering results, i.e., $R_k$, based on the current graph ${\cal{G}}_k$,\\
		Transferring current results $R_{k}$ to multiple views, and calculating new data samples as new input data by averaging data samples in same cluster.\\	
	}
\end{algorithm}

\subsection{Nearest Neighbor Agglomeration}
According to the learned essential cosine distance matrix in CDI, a graph $\cal{G}$, in which only each data sample and its nearest neighbor can be connected, is constructed for clustering. To this end, we have the following formulation:
\begin{equation}
{\cal{G}}(a,b) = \left\{ \begin{array}{l}
1,~{\rm{if}~a = \rm{nearest}(b)~\rm{or}~b = \rm{nearest}(a)},\\
0,~{\rm{otherwise}}.
\end{array} \right.
\end{equation}
where $\rm{nearest}(a)$ is a function to find the nearest neighbor of data sample $a$. Based on the intuition that a sample with its nearest neighbor should be partitioned into the same cluster, we agglomerate connected data samples and get clustering results of this iteration. The first neighbors can be searched effectively via fast approximate nearest neighbor methods, i.e., k-d tree.

\begin{algorithm}[!htbp]
	\caption{MHC with a fixed number of clusters}
	\label{Algorithm2}
	\LinesNumbered 
	\KwIn{Multi-view data, i.e., $\{ {{\bm{X}}^{(i)}}\} _{i = 1}^v$, the required number of clusters $m$, and the closest division clustering results $R_{\rm{closest}}$, which has $m_{\rm{closest}}$ clusters. }
	\KwOut{Clustering results with the required number of clusters $R_{\rm{req}}$.}
	\textbf{Clustering Process:}\\
	\tcp{Initializing}
	Calculating new data samples of multiple views as new input data by averaging samples in the same cluster according to $R_{\rm{closest}}$.\\
	\tcp{Getting the desired results}
	\For{${\rm{iteration}}=m_{\rm{closest}}-m$}{
		\tcp{Performing the CDI}
		\For{$i=1:v$}{
			Calculating distance matrix ${\bm{D}}_{\rm{iteration}}^{(i)}$ based on new input data by Eq.~(\ref{CosineDistance_Xi}),\\
		}
		Getting essential distance matrix ${\bm{D}}_{\rm{iteration}}^{*}$ by Eq.~(\ref{Get_Intrinsic_Distance_Matrix}).\\
		\tcp{Merging two closest clusters}
		Getting new clustering results by merging two closest clusters based on ${\bm{D}}_{\rm{iteration}}^{*}$,
	}
\end{algorithm}

Up to now, clustering results achieved in this iteration are in latent representation. Subsequently, we transfer clustering results from latent representation to multiple views directly. Therefore, the same partitioning, namely, the same new clusters in different views, are achieved for all views. We treat these new clusters as the input data samples for CDI of the next iteration, and each new data sample can be obtained by averaging data samples in the same cluster. It is notable that the agglomeration strategy used in NNA is more reliable and lower in compute complexity compared to most traditional hierarchical clustering algorithms, which use single-linkage, average-linkage, or complete-linkage for agglomeration. 

\subsection{Algorithm of MHC}
The proposed method provides clustering results at different levels of granularity. Algorithm~\ref{Algorithm1} summarizes the proposed method. To be brief, all data samples are treated as separate clusters at the beginning. Then the CDI and NNA are recursively performed until all data samples are clustered into one cluster.

Clearly, clustering results obtained by Algorithm~\ref{Algorithm1} are at an increasingly finer granularity. In some practical applications, such as user segmentation on social networks or online shopping platforms, the exact number of clusters is unknown in most cases. Clustering results at different levels of granularity are significant for these applications. Furthermore, our method can also get results with fixed number of clusters.

To achieve clustering results with the desired number of clusters, we refine the closest division by merging two closest clusters at each iteration, which is similar to the AGNES (AGlomerative NESting) \cite{kaufman1990finding}. However, AGNES is designed for single-view data and uses the single-linkage, average-linkage, or complete-linkage for agglomeration. The method of obtaining the desired number of clusters is summarized in Algorithm~\ref{Algorithm2}.

For the sake of clarity, we take multi-view data, depicted in Fig.~\ref{Flowchart_MHC}, for example. In the $1$-st view, data samples, i.e., $\{ {\bm{x}}_1^{(1)}, {\bm{x}}_2^{(1)}, \cdots ,{\bm{x}}_7^{(1)}\}$, are regraded as $7$ separated clusters, i.e., $\{ {\bm{C}}_1^{(1)}, {\bm{C}}_2^{(1)}, \cdots ,{\bm{C}}_7^{(1)}\}$, at the beginning. According to Eq.~(\ref{CosineDistance_Xi}), we can get the cosine distance of the $1$-st view, i.e., ${\bm{D}}^{(1)}$. The cosine distance of the $2$-nd view, i.e., ${\bm{D}}^{(2)}$ can be achieved similarly. By performing the CDI, the essential cosine distance matrix of latent representation, i.e., ${\bm{D}}^*$ can be calculated by Eq.~(\ref{Get_Intrinsic_Distance_Matrix}). By performing the NNA, without loss of generality, we assume that we find $d_{12}^*$, $d_{21}^*$, $d_{34}^*$, $d_{43}^*$, $d_{56}^*$, and $d_{65}^*$ are equal to $1$ in ${\bm{D}}^*$, therefore, we get clustering results of the 1-st iteration by merging samples $1$ and $2$ into a new cluster, merging samples $3$ and $4$ into a new cluster, merging samples $5$ and $6$ into a new cluster. Then, we transfer current clustering results in latent representation to the $1$-th view and $2$-nd view. For two views, the new clustering results, i.e., $R_1=\{{\bm{C}}_1, {\bm{C}}_2, {\bm{C}}_3\}$, are achieved in the $1$-st iteration. Subsequently, we average data samples in the same cluster as new data samples of these clusters to begin the next iteration. By repeating the CDI and NNA, we attain hierarchical clustering results ${\cal{R}} = \{R_0, R_1, R_2\}$, $R_0$ has $7$ clusters, $R_1$ has $3$ clusters, and $R_2$ has $1$ clusters. To obtain desired clustering results with $2$ clusters, we take multi-view data, $R_1$, and $m=2$ as inputs of Algorithm~\ref{Algorithm2}, and exact clustering results are achieved finally.

\subsection{Model Analysis}
It can be observed that the proposed MHC is free of parameter selection. Given multi-view data, hierarchical clustering results at different levels of granularity are provided in Algorithm~\ref{Algorithm1} without any other inputs. Clearly, it is of significance for the application of multi-view clustering in practice. If the clustering results with a fixed number of clusters is required, we can perform Algorithm~\ref{Algorithm2} based on the hierarchical clustering results obtained in Algorithm~\ref{Algorithm1} to achieve this goal.

Regarding to the computational complexity, the proposed MHC is ${\cal O}(n{\rm{log}}n)$, where $n$ is the number of data samples. Specifically, in the first iteration of Algorithm~\ref{Algorithm1}, cosine distance matrices of multiple views can be pre-computed outside of the main procedure of our MHC. Different from performing the single-linkage, average-linkage, or complete-linkage, we leverage the mean vectors for the distance computation, therefore, the computational burden can be reduced considerably during clustering process.

\section{Experiments}
To validate the superiority of our MHC, experiments are carried out on nine real-world multi-view datasets compared to eleven state-of-the-arts. Experimental results and analyses are provided in this section. All codes are conducted on the workstation with a sixteen-core 2.10 GHz processor and 128 GB of memory.

\subsection{Experimental Settings}
The following datasets are leveraged in this section, including 100leaves\footnote{https://archive.ics.uci.edu/ml/datasets/One-hundred+plant+species+leaves+data+set}, BBCSport\footnote{http://mlg.ucd.ie/datasets/bbc.html}, Caltech101\footnote{http://www.vision.caltech.edu/Image Datasets/Caltech101/}, Football\footnote{http://mlg.ucd.ie/aggregation/}, NottingHill Face \cite{zhang2009character}, Olympics\footnote{http://mlg.ucd.ie/aggregation/}, ORL\footnote{https://www.cl.cam.ac.uk/research/dtg/attarchive/facedatabase.html}, Politicsie\footnote{http://mlg.ucd.ie/aggregation/}, and UCI \cite{asuncion2007uci}. To be specific, statistics of these datasets are summarized in Table~\ref{dataset_information}. 

\begin{table}
	\caption{Statistics of multi-view datasets.}\smallskip
	\label{dataset_information}
	\centering
	\resizebox{0.55\columnwidth}{!}{
		\begin{tabular}{l|c|c|c}
			\toprule
			Datasets & \# Views & \# Samples & \# Clusters\\
			\midrule
			100leaves & 3 & 1600 & 100 \\
			\midrule
			BBCSport & 2 & 544 & 5  \\
			\midrule
			Caltech101 & 6 & 9144 & 102   \\
			\midrule
			Football & 9 & 248 & 20 \\
			\midrule
			NottingHill & 3 & 4660 & 5  \\
			\midrule
			Olympics & 9 & 464 & 28 \\
			\midrule
			ORL & 3 & 400 & 40  \\
			\midrule
			Politicsie & 9 & 348 & 7 \\
			\midrule
			UCI & 3 & 2000 & 10 \\
			\bottomrule
	\end{tabular}}
\end{table}

We compare the proposed approach with eleven state-of-the-art clustering approaches, including, the k-means, low-rank representation (LRR) \cite{liu2013robust}, robust multi-view spectral clustering (RMSC) \cite{xia2014robust}, low-rank tensor constrained multi-view subspace clustering (LT-MSC) \cite{zhang2015low}, auto-weighted multiple graph learning (AMGL) \cite{nie2016parameter}, latent multi-view subspace clustering (LMSC) \cite{zhang2017latent}, binary multi-view clustering (BMVC) \cite{zhang2018binary}, graph-based multi-view clustering (GMC) \cite{wang2019gmc}, multi-view clustering without parameter selection (COMIC) \cite{peng2019comic}, and large-scale multi-view subspace clustering (LMVSC) \cite{kang2020large}. For COMIC, k-means is used to get results with a desired number of clusters. To be clear, k-means and LRR are single-view clustering algorithms, we report best clustering results of multiple views. 

\begin{table*}
	\caption{Clustering results in the metric of ACC.}\smallskip
	\label{ACC_results}
	\centering
	\resizebox{1\textwidth}{!}{
		\begin{tabular}{l|ccccccccc}
			\toprule
			ACC & 100leaves & BBCSport & Caltech101 & Football & NottingHill & Olympics & ORL & Politicsie & UCI\\
			\midrule
			k-means & 0.617 & 0.408 & 0.258 & 0.651 & 0.822 & 0.569 & 0.729 & 0.621 & 0.732  \\
			\midrule
			LRR (TPAMI'13) & 0.462 & 0.797 & 0.241 & 0.834 & 0.794 & 0.780 & 0.812 & 0.681 & 0.871 \\
			\midrule
			RMSC (AAAI'14) & 0.775 & 0.856 & 0.173 & 0.715 & 0.807 & 0.637 & 0.704 & 0.410 & 0.915  \\
			\midrule
			LT-MSC (ICCV'15) & 0.729 & \underline{0.926} & 0.267 & 0.870 & \underline{0.868} & \underline{0.865} & \underline{0.821} & \underline{0.888} & 0.803 \\
			\midrule
			AMGL (IJCAI'16)  & 0.727 & 0.919 & 0.238 & 0.744 & 0.358 & 0.689 & 0.725 & 0.816 & 0.764\\
			\midrule
			LMSC (CVPR'17)  & 0.748 & 0.918 & 0.251 & 0.795 & 0.816 & 0.804 & 0.819 & 0.690 & 0.859\\
			\midrule
			BMVC (TPAMI'18)  & 0.776 & 0.774 & \underline{0.288} & 0.685 & 0.319 & 0.724 & 0.593 & 0.721 & 0.783 \\
			\midrule
			GMC (TKDE'19)  & \underline{0.824} & 0.739 & 0.195 & \underline{0.883} & 0.312 & 0.819 & 0.635 & 0.856 & 0.733 \\
			\midrule
			COMIC (ICML'19)  & 0.407 & 0.793 & 0.111 & 0.772 & 0.707 & 0.759 & 0.567 & 0.713 & \underline{0.940}\\
			\midrule
			LVMSC (AAAI'20)  & 0.561 & 0.605 & 0.112 & 0.657 & 0.726 & 0.631 & 0.555 & 0.569 & 0.714\\
			\midrule
			MHC  & \bf{0.868} & \bf{0.965} & \bf{0.378} & \bf{0.919} & \bf{0.970} & \bf{0.946} & \bf{0.858} & \bf{0.891} & \bf{0.958}\\
			\bottomrule
	\end{tabular}}
\end{table*}

\begin{table*}
	\caption{Clustering results in the metric of NMI.}\smallskip
	\label{NMI_results}
	\centering
	\resizebox{1\textwidth}{!}{
		\begin{tabular}{l|ccccccccc}
			\toprule
			NMI & 100leaves & BBCSport & Caltech101 & Football & NottingHill & Olympics & ORL & Politicsie & UCI\\
			\midrule
			k-means & 0.818 & 0.136 & 0.496 & 0.717 & 0.723 & 0.713 & 0.876 & 0.623 & 0.738  \\
			\midrule
			LRR (TPAMI'13) & 0.736 & 0.700 & 0.495 & 0.850 & 0.579 & 0.867 & \underline{0.924} & 0.779 & 0.768 \\
			\midrule
			RMSC (AAAI'14) & 0.919 & 0.812 & 0.395 & 0.779 & 0.585 & 0.760 & 0.854 & 0.053 & 0.822 \\
			\midrule
			LT-MSC (ICCV'15) & 0.868 & 0.803 & \bf{0.512} & 0.891 & \underline{0.779} & \underline{0.940} & 0.921 & \underline{0.820} & 0.775\\
			\midrule
			AMGL (IJCAI'16)  & 0.890 & \underline{0.864} & 0.391 & 0.802 & 0.129 & 0.810 & 0.883 & 0.764 & 0.798\\
			\midrule
			LMSC (CVPR'17)  & 0.877 & 0.839 & 0.485 & 0.840 & 0.697 & 0.891 & 0.921 & 0.684 & 0.782\\
			\midrule
			BMVC (TPAMI'18)  & 0.909 & 0.557 & 0.505 & 0.749 & 0.071 & 0.807 & 0.746 & 0.636 & 0.796 \\
			\midrule
			GMC (TKDE'19)  & \underline{0.929} & 0.795 & 0.345 & \underline{0.879} & 0.092 & 0.875 & 0.857 & 0.753 & 0.812 \\
			\midrule
			COMIC (ICML'19)  & 0.700 & 0.701 & 0.288 & 0.797 & 0.695 & 0.832 & 0.785 & 0.720 & \underline{0.892}\\
			\midrule
			LVMSC (AAAI'20) & 0.785 & 0.436 & 0.263 & 0.708 & 0.681 & 0.745 & 0.789 & 0.501 & 0.712\\
			\midrule
			MHC  & \bf{0.950} & \bf{0.890} & \underline{0.509} & \bf{0.909} & \bf{0.926} & \bf{0.947} & \bf{0.953} & \bf{0.857} & \bf{0.916}\\
			\bottomrule
	\end{tabular}}
\end{table*}

\begin{table*}
	\caption{Clustering results in the metric of F-measure.}\smallskip
	\label{Fmeasure_results}
	\centering
	\resizebox{1\textwidth}{!}{
		\begin{tabular}{l|ccccccccc}
			\toprule
			F-measure & 100leaves & BBCSport & Caltech101 & Football & NottingHill & Olympics & ORL & Politicsie & UCI\\
			\midrule
			k-means & 0.505 & 0.409 & 0.224 & 0.452 & 0.753 & 0.302 & 0.645 & 0.542 & 0.681  \\
			\midrule
			LRR (TPAMI'13) & 0.352 & 0.761 & 0.179 & 0.745 & 0.653 & 0.710 & 0.765 & 0.676 & 0.763 \\
			\midrule
			RMSC (AAAI'14) & 0.730 & 0.851 & 0.136 & 0.623 & 0.603 & 0.572 & 0.623 & 0.398 & 0.811   \\
			\midrule
			LT-MSC (ICCV'15) & 0.636 & 0.858 & 0.197 & \underline{0.800} & \underline{0.825} & \underline{0.844} & \underline{0.767} & \bf{0.907} & 0.753\\
			\midrule
			AMGL (IJCAI'16)  & 0.410 & \underline{0.901} & 0.066 & 0.610 & 0.369 & 0.584 & 0.535 & 0.776 & 0.722\\
			\midrule
			LMSC (CVPR'17)  & 0.675 & 0.900 & 0.192 & 0.704 & 0.761 & 0.787 & 0.762 & 0.652 & 0.764\\
			\midrule
			BMVC (TPAMI'18) & \underline{0.717} & 0.665 & \underline{0.237} & 0.579 & 0.254 & 0.699 & 0.431 & 0.634 & 0.750 \\
			\midrule
			GMC (TKDE'19) & 0.502 & 0.721 & 0.050 & 0.709 & 0.369 & 0.701 & 0.360 & 0.778 & 0.708 \\
			\midrule
			COMIC (ICML'19) & 0.282 & 0.766 & 0.070 & 0.636 & 0.692 & 0.671 & 0.408 & 0.694 & \underline{0.888}\\
			\midrule
			LVMSC (AAAI'20) & 0.448 & 0.478 & 0.064 & 0.420 & 0.696 & 0.357 & 0.439 & 0.450 & 0.653\\
			\midrule
			MHC  & \bf{0.826} & \bf{0.931} & \bf{0.425} & \bf{0.843} & \bf{0.952} & \bf{0.928} & \bf{0.809} & \underline{0.883} & \bf{0.918}\\
			\bottomrule
	\end{tabular}}
\end{table*}

To measure the clustering quality of different approaches comprehensively, three widely-used criterias are adopted for evaluation metrics, including ACCuarcy (ACC), Normalized Mutual Information (NMI), and F-measure \cite{cao2015diversity,zhang2017latent}. We highlight and underline the best and second best clustering results.

\subsection{Experimental Results and Analysis}
Clustering results are reported in Table~\ref{ACC_results},~\ref{NMI_results},~and~\ref{Fmeasure_results}. In a big picture, the proposed MHC achieves better clustering results than other methods. Compared to single-view methods, our method achieved significant improvements on all multi-view datasets of all metrics, since the complementary information of multiple views can be fully explored in our method. Compared to other multi-view clustering algorithms, remarkable process is also made by the proposed MHC. Taking results on BBCSport, NottingHill Face and UCI for examples, results of our MHC are about $3.9\%$, $10.2\%$, and $1.8\%$ higher than the second best clustering results in the metric of ACC, about $2.6\%$, $14.7\%$, and $2.4\%$ higher than the second best clustering results in the metric of NMI, about $3.0\%$, $12.7\%$, and $3.0\%$ higher than the second best clustering results in the metric of F-measure.

\subsubsection{Clustering results at different levels of granularity.}
To further investigate the proposed MHC, we present clustering results at different levels of granularity, which can be gained by Algorithm~\ref{Algorithm1}. Table~\ref{Clustering_results_granularity} lists the number of clusters achieved at different levels of granularity. It can be observed that clustering results gained by Algorithm~\ref{Algorithm1} can provide the desired or very close number of clusters for most datasets. Besides, Table~\ref{NMI_at_Closed} gives clustering results with the closest number of clusters in the metric of NMI.

Actually, for many real-world applications, the true number of clusters is hard to obtain. Therefore, clustering results at multiple levels of granularity provided by Algorithm~\ref{Algorithm1} is meaningful in practice.

\begin{table}
	\caption{Clustering results at different levels of granularity.}\smallskip
	\label{Clustering_results_granularity}
	\centering
	\resizebox{0.6\columnwidth}{!}{
		\begin{tabular}{l|c|l}
			\toprule
			Datasets & True \# Clusters & \# Clusters in Algorithm~\ref{Algorithm1} \\
			\midrule
			100leaves & 100  & $\{382, 114, 33, 10, 3, 1\}$\\
			\midrule
			BBCSport & 5  & $\{105, 25, 5, 2, 1\}$\\
			\midrule
			Caltech101 & 102 & $\{1657, 259, 56, 14, 5, 1\}$ \\
			\midrule
			Football & 20 & $\{48, 17, 3, 1\}$ \\
			\midrule
			NottingHill & 5 & $\{1282, 339, 104, 29, 9, 3, 1\}$ \\
			\midrule
			Olympics & 28 & $\{106, 29, 11, 2, 1\}$ \\
			\midrule
			ORL & 40 & $\{122, 42, 5, 1\}$ \\
			\midrule
			Politicsie & 7 & $\{62, 13, 3, 1\}$ \\
			\midrule
			UCI & 10 & $\{464, 109, 27, 9, 4, 1\}$ \\
			\bottomrule
	\end{tabular}}
\end{table}

\begin{table}
	\caption{Clustering results with the closed number of clusters in the metric of NMI.}\smallskip
	\label{NMI_at_Closed}
	\centering
	\resizebox{0.6\columnwidth}{!}{
		\begin{tabular}{l|c|c|c}
			\toprule
			Datasets & True \# Clusters & Closed \# Clusters & NMI \\
			\midrule
			100leaves & 100  & 114 & 0.647\\
			\midrule
			BBCSport & 5  & 5 & 0.890\\
			\midrule
			Caltech101 & 102 & 56 & 0.473 \\
			\midrule
			Football & 20 & 17 & 0.855 \\
			\midrule
			NottingHill & 5 & 3 & 0.721 \\
			\midrule
			Olympics & 28 & 29 & 0.803 \\
			\midrule
			ORL & 40 & 42 & 0.957 \\
			\midrule
			Politicsie & 7 & 3 & 0.591 \\
			\midrule
			UCI & 10 & 9 & 0.900 \\
			\bottomrule
	\end{tabular}}
\end{table}

\subsubsection{Running time.}
To show the effectiveness of our method, we compare running time of the proposed MHC to BMVC and LMVSC , which are designed for the large-scale dataset. The computational complexity of BMVC, LMVSC, and MHC are ${\cal O}(n)$, ${\cal O}(n)$, and ${\cal O}(n{\rm{log}}n)$, where $n$ is the number of data samples. It can be observed in Table~\ref{Times} that the proposed MHC can obtain clustering results within 61 seconds for all datasets, including Caltech101 and Notting Hill Face. In general, although BMVC is slightly better in running time, our approach can achieve better clustering results without the parameter selection. Since more iterations are involved in LMVSC to achieve convergence, the running time of LMVSC is slightly larger than the proposed MHC.

\begin{table}
	\caption{Running time (second) of BMVC, LVMSC and the proposed MHC.}\smallskip
	\label{Times}
	\centering
	\resizebox{0.6\columnwidth}{!}{
		\begin{tabular}{l|c|c|c}
			\toprule
			Datasets & ~~BMVC~~ & ~~LVMSC~~ & ~~MHC~~ \\
			\midrule
			100leaves & 0.665 & 9.417 & 1.220 \\
			\midrule
			BBCSport & 0.082 & 1.385 & 0.548 \\
			\midrule
			Caltech101 & 8.157 & 139.75 & 26.628 \\
			\midrule
			Football & 0.061 & 2.306 & 1.007 \\
			\midrule
			NottingHill & 2.532 & 9.08 & 60.431 \\
			\midrule
			Olympics & 0.120 & 4.433 & 2.982 \\
			\midrule
			ORL & 0.080 & 3.668 & 1.079 \\
			\midrule
			Politicsie & 0.076 & 0.611 & 0.902 \\
			\midrule
			UCI & 0.659 & 2.799 & 0.377 \\
			\bottomrule
	\end{tabular}}
\end{table}

\section{Conclusion}
This paper proposes a novel multi-view hierarchical clustering, which can efficiently achieve clustering without parameter selection. By introducing the cosine distance integration and the nearest neighbor agglomeration, the complementary information of multiple view can be fully explored for clustering. Extensive experiments conducted on nine real-world datasets illustrate the superior performance of the proposed method.

\section*{Acknowledgements}

This work is supported by the National Natural Science
Foundation of China under Grant No. 61573273, and the Fundamental Research Funds for Central Universities under Grant No. xzy022020050.

\bibliography{mybibfile}

\end{document}